# Construction and application of artificial intelligence crowdsourcing map based on multi-track GPS data

Yong Wang[1,*]
Information Technology
University of Aberdeen
Aberdeen, United Kingdom
fredia4jane@gmail.com

Yanlin Zhou[1]
Computer Science
Johns Hopkins University
MD,USA
popojoyzhou@gmail.com

Huan Ji[2]
Information Science
MSEM in Trine University
San jose, CA, 95133
hji23@my.trine.edu

Zheng He[3]
Applied Analytics
Columbia University
NY,USA
z2541@columbia.edu

Xinyu Shen[4]
Biostatistics
Columbia University
TX,USA
xshe1007@gmail.com Frisco

*Abstract*—In recent years, the rapid development of high-precision map technology combined with artificial intelligence has ushered in a new development opportunity in the field of intelligent vehicles. High-precision map technology is an important guarantee for intelligent vehicles to achieve autonomous driving. However, due to the lack of research on high-precision map technology, it is difficult to rationally use this technology in the field of intelligent vehicles. Therefore, relevant researchers studied a fast and effective algorithm to generate high-precision GPS data from a large number of low-precision GPS trajectory data fusion, and generated several key data points to simplify the description of GPS trajectory, and realized the "crowdsourced update" model based on a large number of social vehicles for map data collection came into being. This kind of algorithm has the important significance to improve the data accuracy, reduce the measurement cost and reduce the data storage space. On this basis, this paper analyzes the implementation form of crowdsourcing map, so as to improve the various information data in the high-precision map according to the actual situation, and promote the high-precision map can be reasonably applied to the intelligent car.

## I. Introduction (Heading 1)

While the digital map is used more and more widely, the accuracy of the digital map is constantly improved, and the tracking and positioning based on the accurate digital map plays an increasingly important role. At present, in train location, the odometer, query transponder and interrail cable are used to ensure that the train location accuracy is within the design specification min value[1]. On this basis, adding satellite positioning data can improve the reliability of train positioning and further improve the positioning accuracy. Using the Global Positioning System (GPS) digital map system to track and position the traffic system in real time can effectively improve the positioning performance, reduce the operating cost, and improve the operating efficiency. In the positioning based on GPS digital map, it is of great practical significance to improve the accuracy of digital map.

In terms of map generation, it is one of the main ways to generate high-precision track map based on GPS measurement data. In this paper, the error data detection algorithm of GPS data points is proposed, and a nonlinear combined GPS data point reduction model is presented. According to the literature, it can be seen that in the process of extracting GPS track data of social cars and adding it to the crowdsourced high-precision map, three constrained main curve algorithms can be adopted. On the basis of the algorithm, many crowdsourced map researches use V-SLAM as the basis of car-end mapping through artificial intelligence convolutional algorithm[2]. HERB et al. used ORB-SLAM as a local map at the end of the vehicle to reconstruct the algorithm.

Convolutional Neural Networks (CNNS) are used to segment the semantics of the key frames and upload the feature information, and then the features of each key frame are matched and aligned in the cloud. Finally, the uploaded feature information is used to merge and reconstruct the road edge.

---

[1] These authors contributed equally to this work and should be considered co-first authors

## II. RELATED WORK

### A. Highly refined crowdsourced maps

High-precision crowdsourcing maps provide rich road environment information. Compared with traditional navigation electronic maps, they can be compared from the following dimensions, and the differences are described below[3].

1) Data accuracy: High precision is the biggest feature of high-precision maps. The accuracy of high-precision maps is at the sub-meter or even centimeter level, while the accuracy of traditional navigation electronic maps is at the meter level.

2) Data dimension: high-precision map data has more abundant elements, including almost all road elements visible in the process of vehicle driving. In addition to navigation map data, it also contains rich geometric and attribute information such as lanes, landmarks, protective facilities, raising facilities, road edges, traffic signs, traffic lights, poles, intersections and stop lines. Traditional navigation electronic map data only records road level data information.

3) Data freshness: In order to ensure the safety and efficiency of automatic driving, it is necessary to provide accurate road environment information for automatic driving vehicles, and the data freshness of high-precision maps has higher requirements. Road construction, road opening, traffic sign changes, etc. Traffic network information is changing every day, and these changes need to be reflected on the high-precision map in time.[4]

4) Application objective: High-precision maps are used in autonomous driving applications to assist them in vehicle planning and control to ensure driving safety, while traditional navigation electronic maps are for human drivers to provide them with convenient navigation services.

### B. Multi-track GPS data acquisition

1) Field collection: Before field collection, the collection scope, collection content and objectives will be defined. Select a proper collection device based on the collection requirements and ensure that the device works properly and has sufficient data storage space (hard disks). After the preparation work is completed, the field data is collected according to the planned collection area. High-precision map acquisition vehicles usually need to be equipped with a series of professional equipment to collect map information[5-8]: DGNSS receiver, receiving signals from satellite systems (such as GPS.GLONASS, Galileo, etc.), to obtain accurate position information of the vehicle; By transmitting laser beams to measure the return time, 3D point cloud data of the surrounding environment can be quickly obtained, so as to obtain accurate distance and shape information of road elements, surrounding buildings and other ground objects.

2) Data processing: This process is mainly responsible for converting the collected and returned tracks, point clouds, photos and other data into usable data for high-precision map production. It is a pre-module for high-precision map production, mainly including data resolution, segmentation, alignment recognition and other links. Among them, the automatic recognition algorithm can output vector data such as lane lines, road signs and ground signs based on point cloud and photo recognition, which can be used as a data source for manual work and greatly improve data production efficiency.

3) Data production: The results of automatic identification and the original version of the data (if not the first production) are processed by difference fusion, and the job tasks are assigned to the data production personnel through the whole process scheduling of online tasks. The data producer completes the editing process according to the pre-defined data specifications and processes. Some map elements can be automatically extracted and assigned by batch processing program on the basis of manual work.

4) Data inspection: data specification and logic-related quality inspection and quality control of data production results are carried out to check the correctness of data in each link, which is an important barrier for data quality control, including single-task inspection and multi-task inspection.

5) Conversion and compilation: mainly converts the data after checking no problems into a data format that meets the requirements of automatic driving applications, and is the final product form of high-precision maps. After it is sent to the State Bureau of Surveying and Mapping for review in accordance with the prescribed procedures and the approval number is obtained, it will be distributed to the vehicle through the cloud service for the use of autonomous driving applications

## III. MATHEMATICAL MODELING OF GPS TRAJECTORY DATA

### A. Question raising

In order to obtain the high precision GPS data, it is necessary to fuse the low precision GPS data, the purpose is to replace the discrete points in the GPS digital map with the high precision polyline segment. The broken line segment is composed of a top point and a straight line[9]. All the three algorithms use the broken line segment to represent the curve, that is, the skeleton of the data curve.

Let the vertices of the data be $P_i(i=1,2,3,...n)$, the line is $L_{i-1,i}$ $(i=2,3,4,...n)$, represents the line from the point $i-1$ to $i$. According to the data nearest principle, data points $t_j(j=1,2,3,...,m)$ is divided into adjacent lines.

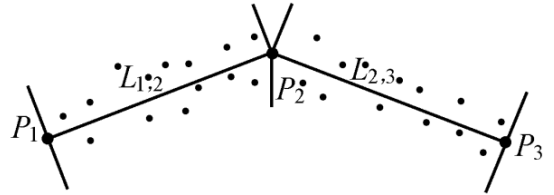

Figure 1. Data recency principle model

### B. Mathematical model building

First, we need to get the distance between the data point and the adjacent line segment. According to the distance formula between the point and the line, we can get the data point $t_j(j=1,2,3...,m)$ The distance from the nearest principle line segment is:

$$D_{line} = |(xp_i - xp_{i-1}) \cdot yt_j - (yp_i - yp_{i-1}) \cdot xt_j + xp_{i-1}$$
$$\cdot yp_i - yp_{i-1} \cdot xp_i|/$$
$$\sqrt{(xp_i - xp_{i-1})^2 + (yp_i - yp_{i-1})^2}$$
$$i=1,2,3\ldots,n \quad j=1,2,3\ldots,m \quad (1)$$

In the formula, $xt_j$ and $yt_j$ are the horizontal and vertical coordinates of data points $t_j$; $xp_i$ and $yp_i$ are the horizontal and vertical coordinates of $P_i$.

### C. Multi-GPS trajectory data fusion algorithm

Multi-gps trajectory data fusion algorithm needs to divide the generation of trajectory into initial trajectory and final trajectory, which is complicated and difficult to realize. In this paper, we propose a Maximum Distance Algorithm (MDA)[10-11], which does not need to divide trajectories into initial and final trajectories, but only needs to locally optimize the newly generated vertices. The algorithm assumes that the starting and ending coordinates of the data are fixed and accurate. The steps of MDA algorithm are as follows:

The beginning and end of the connection are the initial line segment Li.

Procedure Step 1 Determine data point t. (j=1,2,··,m) to the initial line

Whether the average projection distance D of Step 2 is less than the set error E? If yes, end; if no, enter

Step 3 Find out the point t with the greatest distance between the data point and the adjacent line segment L, take this point as the center of the circle, and all the data points within a certain distance d<o(the radius of the circle), find the center of gravity point of these data points, and get a new vertex P, connect the new vertex in turn, and then divide the data into the nearest line segment or vertex region. Determine whether the average projection distance D is less than the set error E. If not, repeat step 3 and continue to add vertices. If the condition is met, the generation of line segments ends.

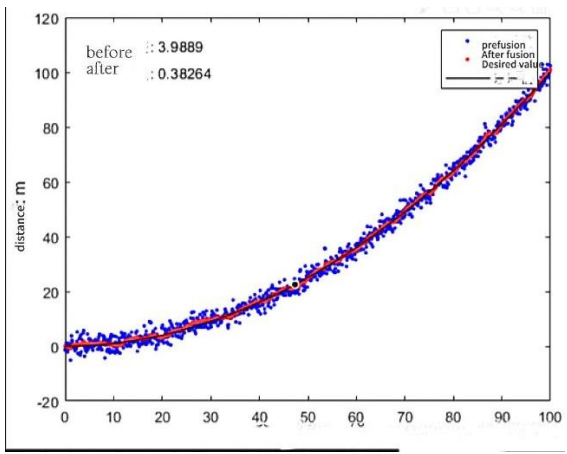

Figure 2.  Diagram of maximum distance data fusion algorithm

According to certain rules, the large scale GPS data is divided into several smaller sub-data blocks. The steps of the algorithm are as follows:

Step 1 Determine the block range w and step variable s. If the value is <w, go to Step 2

Step 2 Start from the leftmost data and store the data in the range w, sort the data in the y direction from smallest to largest, and determine the difference between adjacent data in the y direction. If the difference is greater than the given error E, store the data from this place, and so on. When you're done, go in

Step 3. Step 3 Process the data stored separately, find the center of gravity of these data in turn, and obtain the new vertex of the data

Step 4 Move the block range to the right by a step length and enter step 2 until all data is processed and the generation of connected vertex line segments is complete.

## IV. ALGORITHM INDEX AND METHODOLOGY

### A. Data index

In this paper, the performance indexes of the following five evaluation algorithms are given to comprehensively evaluate the performance of the three proposed algorithms[12].

(1) Operation time t: that is, the total time used by the program from the start to the end of operation, which can reflect the operation efficiency of the algorithm. The smaller the operation time, the higher the efficiency; otherwise, the lower the efficiency

(2) Storage space i: that is, the algorithm finally uses i data points to fully represent the curve, and the smaller i is, the empty space occupied by data storage the smaller the room.

(3) Data reduction rate r: represents the proportion of storage space in the original total GPS data points $r=(i/n) \cdot 100\%$, where n is the total data points, the smaller the reduction rate means the fewer data points used, and the better the extraction effect of the algorithm.

(4) The average value of the lateral error E: the arithmetic average of the projected distance between all points and the corresponding curve segment. The smaller the value, the smaller the error and the better the robustness of the algorithm, which is the basis for judging the error size of the algorithm

(5) Maximum transverse error Emax: the maximum projection distance of all points to the corresponding curve segment, which can test whether the algorithm meets the requirements of horizontal error, and also indirectly indicates the robustness of the algorithm. The smaller the Emax, the better.

### B. GPS data experiment case

In order to analyze and compare the algorithms, the three algorithms were respectively applied to the simulation experimental data[13]. The data source was the 10 dynamic measurements of the actual GPS receiver on the playground of Beijing Jiao tong University (which had been converted into coordinates in the rectangular coordinate system, unit m), with a total of 1599 data points.

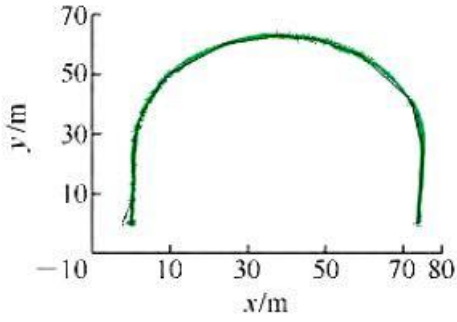

Figure 3.  MDA algorithm processes simulation experiment data

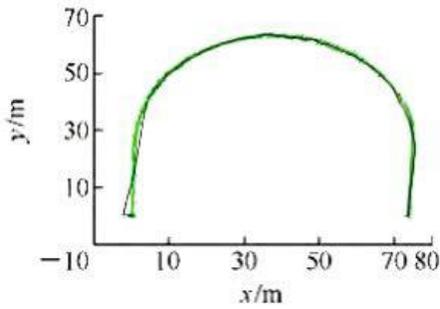

Figure 4.  DPA algorithm processes simulation experiment data

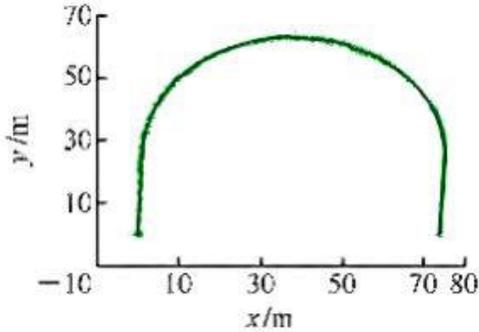

Figure 5.  ARA algorithm processes simulation experiment data

According to the simulation results of the algorithm, except that some line segments of the DPA algorithm are not very integrated with the original data, its capital is well matched with the original data, and the average error is within 1 m, and the result is good[12-13]. The MDA algorithm has the longest time and the accuracy is in the middle of the three. After getting a new vertex, the algorithm has to sort the data around the vertex in the most recent principle, which consumes a lot of time. DPA algorithm has the fastest operation time and the worst accuracy. It sacrifices part of the accuracy to improve the operation efficiency. The error is large, and the obtained curve deviates far from the original GPS data point. The time of ARA algorithm is much smaller than that of MDA algorithm, the storage space is slightly larger, the curve segment obtained is the smoothest, the error is also the smallest among the three algorithms, and the accuracy is the highest.

*C. Highly refined crowdsourced map data structure*

First, the road network. The basic roads in traditional 2D navigation electronic maps are usually used as the main reference. Appropriate addition of three-dimensional information to ensure that the final road geometry is more accurate. In addition to various types of roads and intersections, it should also include the relationship between roads and each lane, which is mainly represented by foreign key correlation in the physical data model.

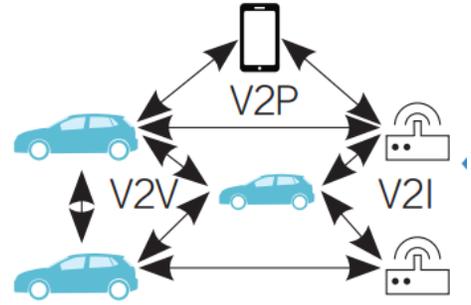

Figure 6.  Crowdsource basic data acquisition and processing nodes

Second, the lane network. The main purpose is to elaborate the geometric position information of the lane. Under normal circumstances, the main components of the lane should include the corresponding lane reference line and connection points and other relevant content. On the other hand, the lanes of each section should form a direct relationship through more reasonable connection points. Nowadays, the lane network has become an important benchmark for the macro-lane level path optimization work, and it is also the car.

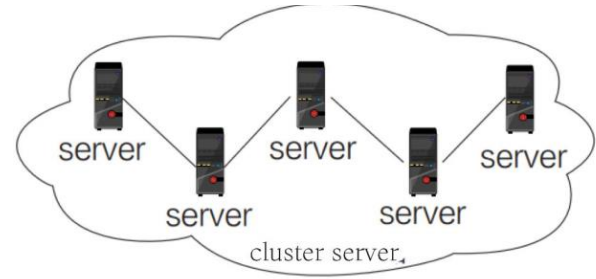

Figure 7.  Intelligent high-precision map data service center

In order to promote intelligent high-precision maps to maintain periodic data updates for a long time, it should be necessary to obtain more road data in the corresponding range as the main support[14]. Only relying on a single surveying and mapping operation can not fully realize the full coverage of the operation in the first time. In general, user crowdsourcing, as an important way to ensure that high-precision maps meet real-time standards, is also a major component of the final form of high-precision maps. Nowadays, crowdsourcing has become a unique distribution and execution mechanism for group intelligence perception to implement distributed tasks[15]. It can not only be reasonably allocated to individuals to complete tasks according to various needs, including regional data processing and data visualization, but also facilitate the smooth development of such functions as road condition control and accident warning.

## V. Conclusions

All in all, intelligent high-precision maps have become an important part of the future travel of the masses of people. Artificial intelligence combined with multi-track GPS data has significantly boosted the development of crowdsourced maps. Using intelligent algorithms, this technology can efficiently process and analyze large amounts of GPS track data to generate more accurate and detailed information on the map. This not only improves the quality and practicality of maps, but also greatly promotes the progress of urban planning, traffic management and navigation systems, and makes full use of the advantages of big data decision-making to realize rational allocation of resources from various aspects and angles to make travel safer and more reliable.

Therefore, in the future development, with the continuous progress of artificial intelligence technology and the improvement of GPS data collection capabilities, we can predict that crowdsourced maps will become more intelligent and accurate. Future crowdsourced maps may integrate more types of data, such as real-time traffic information and environmental monitoring data, to provide more comprehensive and in-depth analysis. In addition, as the demand for personalized and context-sensitive services grows, crowdsourced maps are likely to become more personalized and dynamic, better serving the specific needs of users